\pdfoutput=1

\documentclass[11pt]{article}

\usepackage{acl} 

\usepackage{times}
\usepackage{latexsym}

\usepackage[T1]{fontenc}

\usepackage[utf8]{inputenc}

\usepackage{microtype}

\usepackage{inconsolata}
\usepackage{listings}

\title{\testbed{}: Conversation History AwaReness Probing for Knowledge-grounded Dialogue Systems}

\author{Abbas Ghaddar$^{1\spadesuit}$\hspace{3mm}David Alfonso-Hermelo$^1$\hspace{3mm}Philippe Langlais$^2$ \\
{\bf Mehdi Rezagholizadeh$^1$\hspace{3mm}Boxing Chen$^1$\hspace{3mm}Prasanna Parthasarathi$^1$} \\
$^1$ Huawei Noah's Ark Lab\\
$^2$ RALI/DIRO, Universit\'e de Montr\'eal, Canada\\
 \texttt{abbas.ghaddar@huawei.com}\\ 
}

\usepackage{booktabs}
\usepackage{eqparbox}
\usepackage{arydshln}
\usepackage{subcaption}
\usepackage{rotating,csquotes}
\usepackage{xcolor}
\usepackage{multirow}
\usepackage{framed}
\usepackage{booktabs}
\usepackage{amsmath}
\usepackage{amssymb}
\usepackage{mathtools}
\usepackage{upgreek}
\usepackage{soul}
\usepackage{amsfonts}
\usepackage{todonotes}
\usepackage{placeins}
\usepackage[most]{tcolorbox}
\usepackage{tabularx}
\usepackage{spverbatim}

\newtcbox{\mybox}[1][]{enhanced, colframe=blue, colback=blue!15, 
	frame style={opacity=0.25}, interior style={opacity=0.25}, 
	nobeforeafter, tcbox raise base, shrink tight, extrude by=1mm, #1}

\newcommand{\bert}{\textsc{Bert}}

\newcommand{\testbed}{\textsc{CHARP}}
\newcommand{\etestbed}{e\textsc{CHARP}}
\newcommand{\htestbed}{h\textsc{CHARP}}
\newcommand{\critic}{\textsc{Critic}}

\newcommand{\flan}{\texttt{FLAN}}
\newcommand{\godel}{\texttt{GODEL}}
\newcommand{\llama}{\texttt{Llama-2-7B}}
\newcommand{\mixtral}{\texttt{Mixtral}}
\newcommand{\chatgpt}{\texttt{ChatGPT}}
\newcommand{\gptf}{\texttt{GPT-4}}
\newcommand{\gptfturbo}{\texttt{GPT4-turbo}}

\newcommand{\mightmention}[1]{}
\newcommand{\problem}[1]{\textcolor{red}{$\star$}}
\newcommand{\answer}[1]{\textcolor{blue}{$\#$}}
\newcommand{\todoreview}[1]{\textcolor{green}{$@$}}

\DeclareMathAlphabet\mathbfcal{OMS}{cmsy}{b}{n}

\newcommand{\quotes}[1]{``#1''}
\newcommand{\halIcon}{\includegraphics[height=\baselineskip]{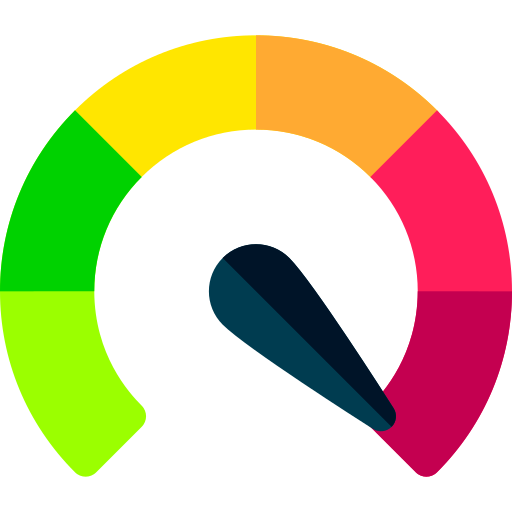}}
\newcommand{\nohalIcon}{\includegraphics[height=\baselineskip]{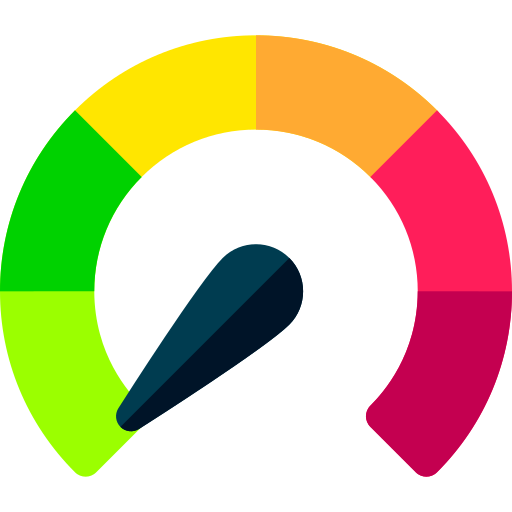}}

\newcommand\blfootnote[1]{%
  \begingroup
  \renewcommand\thefootnote{}\footnote{#1}%
  \addtocounter{footnote}{-1}%
  \endgroup
}

\begin{document}
\maketitle
\begin{abstract}
\blfootnote{$^{\spadesuit}$Corresponding author.}
In this work, we dive deep into one of
the popular knowledge-grounded dialogue benchmarks that focus on faithfulness, FaithDial. We show that a significant portion of the FaithDial data contains annotation artifacts, which may bias models towards completely ignoring the conversation history. We therefore introduce \testbed{}, a diagnostic test set, designed for an improved evaluation of hallucinations in conversational model. \testbed{} not only measures hallucination but also the compliance of the models to the conversation task. Our extensive analysis reveals that models primarily exhibit poor performance on \testbed{} due to their inability to effectively attend to and reason over the conversation history. Furthermore, the evaluation methods of FaithDial fail to capture these shortcomings, neglecting the conversational history. Our findings indicate that there is substantial room for contribution in both dataset creation and hallucination evaluation for knowledge-grounded dialogue, and that \testbed{} can serve as a tool for monitoring the progress in this particular research area. \testbed{} is publicly available at \url{https://huggingface.co/datasets/huawei-noah/CHARP}

\end{abstract}

\section{Introduction}

\begin{figure}[!ht]
    \centering
    \includegraphics[width=1.0\columnwidth]{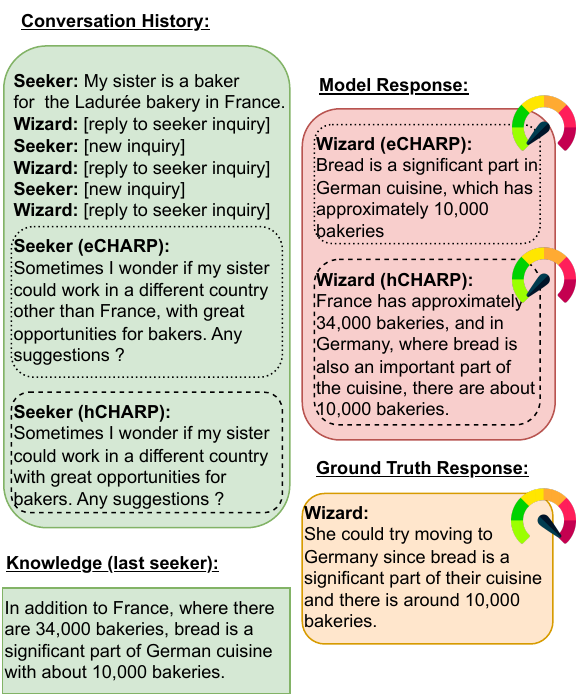}
    \caption{
    \testbed{} consists of 2 subsets, where only the last seeker utterance differs: a self-contained \textit{easy} version (\etestbed{}), and a \textit{hard} (\htestbed{}) which requires reasoning over the conversation history and the provided knowledge that corresponds to the last seeker. In addition to the ground truth response annotation, we show the predictions of a model (\flan{-base}) tuned on the FaithDial training data.\halIcon{} and \nohalIcon{} indicate whether the FaithDial \critic{} labels a response as a hallucination or not. Green boxes indicate model inputs, while pink and orange ones show predicted, and gold responses.}
    \label{fig:charp_illustration}
\end{figure}

Despite the success of general purpose large language models (LLMs) \cite{bommasani2021opportunities}, the utility of the generated texts rests in its relevance and knowledge grounding. The task of information seeking dialogue~\cite{ghazvininejad2018knowledge,lewis2020retrieval} is a touchstone for knowledge grounded generation. The task evaluates a system's ability to respond to user queries while it remains faithful to the knowledge. A system response not adhering to this would be deemed unfaithful. This topic has received considerable attention resulting in several diagnostics and mitigation techniques for texts that lack knowledge grounding and are hallucinatory in nature~\cite{dziri2019evaluating,dziri2022origin,dziri2022evaluating}.

\citet{dziri2022faithdial}'s work in this direction--- FaithDial--- provides a benchmark with hallucination-free annotations, a hallucination detector, and a comprehensive evaluation framework has garnered attention and follow up works \cite{deng-etal-2023-towards,daheim2023elastic}. \citet{dziri2022faithdial} show that T5-base model trained on these annotations restricts hallucination only in $1.4$\% of its responses
, or even $0.3$\% as reported by \cite{daheim2023elastic}. While it appears that hallucination is \emph{under control} at least under the experimental protocol defined in FaithDial, we observe that, though the annotations created in \cite{dziri2022faithdial} are free from hallucinations, they introduce artifacts. These artifacts bias models trained on it to predict the response based solely on the provided knowledge, while ignoring the dialogue history. 

We validate this hypothesis with a controlled evaluation set called \testbed{} (\textbf{C}onversation \textbf{H}istory \textbf{A}wa\textbf{R}eness \textbf{P}robing) with its \emph{easy} and \emph{hard} versions denoted as \etestbed{} and \htestbed{} respectively. The proposed diagnostic set (\S \ref{sec:charp}) not only evaluates hallucinations with respect to the provided knowledge but also its relevance to the conversation history (\autoref{fig:charp_illustration}). \testbed{} is created by annotating on top of $1,080$ samples from FaithDial validation dataset.
\testbed{} tests whether models attend or ignore the history to select the appropriate knowledge when the correct knowledge is augmented with a distracting fact that is irrelevant to the conversation.

Evaluating models using automatic metrics, LLM APIs, and human scorers, we find that training with FaithDial biases the models to ignore conversation history, as assessed with \testbed{}, while remaining faithful to the knowledge (\S \ref{sec:Probing History Awareness}).
Interestingly, this phenomenon was elusive to be observed 
neither with the suite of evaluation methods nor the hallucination detector proposed in \cite{dziri2022faithdial}. Instead, we find the FaithDial detector scoring \testbed{} \emph{gold} responses as hallucinatory ($16.0$\%) that is higher than the hallucination rate ($0.4$\%) of a system performing poorly on \testbed{} as evaluated by a human (\S \ref{sec:charp-auto-eval}).

To understand this, we conduct a thorough human evaluation to identify $6$ different types of errors to be considered in knowledge-grounded response generation. We find human annotation to be effective (\S \ref{sec:Human Evaluation}) in identifying the error types. Further ablations with FaithDial training on human evaluation confirm that the dataset biases the models to look away from the conversation history. We find the evaluation with powerful LLM APIs to be correlated with humans (\S \ref{sec:GPT-4 Evaluation}), and is a better proxy metric for this task over the FaithDial metrics. Overall, this study suggests that despite recent progress reported in hallucination mitigation, developing a model that is simultaneously aware of the conversation history and non-hallucinatory remains an open problem in information-seeking dialogue.

\section{Related Work}

Constructing diagnostic sets with curated adversarial or counterfactual examples, has been shown to be an effective approach across NLP tasks to capture such artifacts that standard evaluation sets fail to detect. For instance, HANS~\cite{mccoy2019right}, FEVER~\cite{schuster2019towards},  PAWS~\cite{zhang2019paws}, CORE~\cite{rosenman2020exposing}, NRB~\cite{ghaddar2021context}, and NATURE~\cite{alfonso2021nature} datasets are vital in identifying biases of models solving tasks like textual entailment, fact verification, paraphrase identification, relation extraction, named entity recognition, and intent detection respectively. 

Studies~\cite{taori2023stanford,chen2023alpagasus,conover2023free} on the recent trend of large-scale pretraining \cite{ouyang2022training,shuster2022blenderbot} show that data quality affects the models in inheriting biases from the artifacts 
embedded in data. 
Especially in information seeking dialogues, \citet{dziri2022evaluating} show that unfaithfulness to the given knowledge is a dominant type of hallucination. \citet{dziri2022origin} show that most information-seeking dialogue datasets like CMU-DoG~\cite{zhou2018dataset}, TopicalChat~\cite{gopalakrishnan2019topical}, and Wizard of Wikipedia (WoW; \citealt{dinan2018wizard}) contain a high ratio of hallucinations, with WoW dataset being the least affected. \citet{dziri2022faithdial} 
propose FaithDial built through replacing hallucinatory WoW annotations with with knowledge faithful responses.

\citet{dziri2022faithdial} show that the models trained on Faithdial show significant reduction in hallucination.
Further studies such as the one from \citet{daheim2023elastic} recently demonstrated that \flan{-T5-base}~\cite{longpre2023flan} can fetch further reduction in hallucination by training on FaithDial over T5-Base~\cite{raffel2019exploring} as reported in \cite{dziri2022faithdial}.
\citet{daheim2023elastic} also propose Elastic Weight Removal (EWR) hallucination mitigation method to reduce the rate of unfaithfulness response generation. 
The authors report high BERTScore similarity~\cite{zhang2019bertscore} between the model response and both the ground truth 
and the provided knowledge
suggesting the faithfulness of the response generated. As training an LM to attend to the knowledge could inadvertently result in ignoring the history of turns leading to poor reasoning of the model, we, in this work, propose \testbed{} that serves as a diagnostic set to measure this phenomenon.

\section{Experimental Setting}
\subsection{Dataset and Task}

We focus on the FaithDial dataset~\cite{dziri2022faithdial} and adhere to its task formulation, where given the history of utterances and a knowledge supplement, a trained model predicts the next response of a \textit{Wizard} bot engaged in a conversation with an information-seeking human (\textit{Seeker}). We assume that the correct knowledge is given, and no retrieval step is performed. A response is considered hallucinatory if it contains information unsupported by the given knowledge snippet.

\subsection{Models and Implementation}
\label{sec:Models and Implementation}

We experiment with the vanilla T5 model~\cite{raffel2019exploring} and two of its derivative variants, namely, Flan-T5~\cite{chung2022scaling} and GODEL~\cite{peng2022godel}. The former was fine-tuned on $1,000$ NLP datasets mapped to an instruction tuning format, while the latter was further pre-trained on $551$M  multi-turn dialogues, and $5$M instruction- and knowledge-grounded dialogues. We primarily focus on the \textit{base} size models, maintaining the same hyperparameters and implementation settings for consistency with previous works~\cite{dziri2022faithdial,daheim2023elastic}. We train all models for a maximum of $20$ epochs and use early stopping based on the validation set performance, and report results on the test set. We use a beam of $5$ during inference in all experiments.

\subsection{Evaluation Metrics}

Following \cite{dziri2022faithdial, daheim2023elastic}, we report the similarity between gold ($y$) and predicted response ($y'$) with BLEU~\cite{papineni2002bleu}, and BERTScore~\cite{zhang2019bertscore}. We measure the hallucination rate using the faithfulness Critic (\critic{}) provided by the FaithDial benchmark\footnote{A detailed description of the development of the FaithDial \critic{} can be found in Appendix~\ref{sec:app:FaithDail Critic}.} and by computing a BERTScore between the knowledge ($k$) and predicted response. 

\subsection{Results Integrity}
\label{sec:app:Results Integrity}

\autoref{tab:baseline_faithdial} is the reproduced test performances of models trained on the FaithDial dataset. For comparable evaluation we consider the baseline models from \cite{dziri2022faithdial}: a vanilla T5-base model and its variant that employs InfoNCE~\cite{oord2018representation} loss for hallucination mitigation; and \cite{daheim2023elastic}: the \flan{-base} model and its variant that utilizes the EWR method for hallucination mitigation. Our reproduced baselines use three different backbone models: \flan{-base}, and the \godel{-base} and \godel{-large} models.

\begin{table}[!htp]
    \begin{center}
    
    \resizebox{\columnwidth}{!}{
\begin{tabular}{lcccc}
\toprule
\multirow{2}{*}{\bf Models} & {\footnotesize \textbf{BLEU}$\uparrow$} & \footnotesize{\textbf{Critic $\downarrow$}} & \multicolumn{2}{c}{\footnotesize \textbf{BERTScore}$\uparrow$} \\
& {{ $(y, y')$}} & {{ $(k, y')$}} & {{ $(y, y')$}} & {{ $(k, y')$}}\\
\midrule
\multicolumn{5}{c}{\textit{ previous works}} \\
\midrule
T5-base $[1]$ & 10.3  & 4.3 & -  & 41.0 \\
\hspace{3mm} +InfoNCE  & 10.9 & 1.4 & - & 39.0 \\
\flan{-base} $[2]$ & 15.1  & 0.3 & 69.6 & 80.9  \\
\hspace{3mm} +EWR & 14.9  & 0.1 & 70.1 & 81.7  \\
\midrule
\multicolumn{5}{c}{\textit{ our re-implementations}} \\
\midrule
\flan{-base} & 15.3 & 0.3  & 69.9  & 80.8 \\
\godel{-base} & 15.5 & 0.3  & 70.2  & 80.5 \\
\godel{-large} & 15.8 & 0.3  & 70.5  & 81.1 \\

\bottomrule
\end{tabular}

    }	
 \end{center}	
	
\caption{Test set performances of previous works alongside our re-implemented baseline models finetuned on the FaithDial dataset. $[1]$ and $[2]$ refer to baselines results directly copied from \cite{dziri2022faithdial} and \cite{daheim2023elastic} respectively. All scores are scaled within the range of $[0, 100]$.}
	\label{tab:baseline_faithdial}
\end{table}

As reported in \citep{daheim2023elastic} the  \flan{-base} models achieve a remarkably low hallucination ratio of only $0.3$\%, significantly improving upon the best FaithDial baselines (both with and without hallucination mitigation methods). Our re-implementation of the \flan{-base} results are similar to \citet{daheim2023elastic}'s based on the significance test on the samples with p-value ($0.58$). This enables a fair comparison of the claims across our experiments 
with \testbed{} and the existing results in the literature. 

In addition, we used the set up to train different models for benchmarking: a dialogue pretrained  (\godel{-base}) and its large version (e.g., \godel{-large}). We observe that these models only hold modest improvements across various metrics.

\subsection{Probing History Awareness}
\label{sec:Probing History Awareness}

While the strong results across models on FaithDial dataset leave no doubts about their faithfulness to the given knowledge, we investigate whether this comes at the expense of an important input component: the conversation history.  To that, we test trained models on truncated conversation history---providing only the last $k$ turns (denoted h=$k$) or no history at all (h=$\emptyset$). As we observed the value of $k$ in the dataset to not affect the performance beyond $3$ despite the average number of turns being $7$, we vary the value of $k$ only in the range of $[0$-$3]$. 
Further, to account for distribution shifts, we fine-tune models on variations of the training data with truncated conversation histories for comparison. 
We used \godel{-base} as the backbone model in this experiment as it shows superior performances compared to \flan{-base} in \autoref{tab:baseline_faithdial}. We benchmark the performance of \godel{-base} on truncated history evaluation in \autoref{tab:history_anal}.

\begin{table}[!ht]
\centering
\resizebox{\columnwidth}{!}{
\begin{tabular}{lcccc}
\toprule
\multirow{2}{*}{} & {\footnotesize \textbf{BLEU}$\uparrow$} & \footnotesize{\textbf{Critic $\downarrow$}} & \multicolumn{2}{c}{\footnotesize \textbf{BERTScore}$\uparrow$} \\
& {{ $(y, y')$}} & {{ $(k, y')$}} & {{ $(y, y')$}} & {{ $(k, y')$}}\\

\midrule
\multicolumn{5}{c}{\textit{\bf eval: h=$all$}} \\
\midrule
train: h=$all$ & 15.5 & 0.3  & 70.2  & 80.5 \\
train: h=$3$  & 15.4      & 0.4           & 70.0       & 79.2        \\
train: h=$2$ & 15.2      & 0.5           & 69.9       & 78.7        \\
train: h=$1$ & 15.0      & 0.6           & 69.8       & 78.2        \\
train: h=$\emptyset$ & 11.9      & 8.4           & 65.8       & 72.8        \\

\midrule
\multicolumn{5}{c}{\textit{\bf eval: h=$3$}} \\
\midrule
train: h=$all$ & 15.3      & 0.4           & 69.9       & 79.9        \\
train: h=$3$ & 15.1      & 0.3           & 69.9       & 80.4 \\
\midrule      
\multicolumn{5}{c}{\textit{\bf eval: h=$2$}} \\
\midrule
train: h=$all$ & 15.1 & 0.3 & 69.9 & 80.6 \\
train: h=$2$ & 15.0 & 0.2 & 69.9 & 80.9 \\
\midrule
 
\multicolumn{5}{c}{\textit{\bf eval: h=$1$}} \\
\midrule    
train: h=$all$ & 14.3 & 0.3 & 69.4 & 81.1 \\
train: h=$1$ & 14.3 & 0.2 & 69.5 & 81.3 \\
\midrule

\multicolumn{5}{c}{\textit{\bf eval: h=$\emptyset$}} \\
\midrule      
train: h=$all$ & 13.1 & 0.0 & 66.6 & 82.1 \\
train: h=$\emptyset$ & 12.7 & 0.0 & 67.1 & 83.9 \\
    
    \bottomrule
    \end{tabular}
    }
    \caption{Performance of \godel{-base} models, trained and evaluated on truncated versions of conversation history from the training and test splits of FaithDial, respectively. Here, h=$i$ means only using the last $i$ turns in the conversion history when training (train:) or evaluating (eval:) models. h=$\emptyset$ and h=$all$ denote using no history and the entire history turns, respectively. All scores are scaled within the range of $[0, 100]$.}
    \label{tab:history_anal}
\end{table}

We notice that the performances on the original test set (eval: h=$all$) of models trained on truncated history (train: $h\!\in\!\{3,2,1\}$) barely drop across metrics. For instance, the hallucination ratio (\critic{} score) slightly increases by 0.1\%  each time the conversational history contains one fewer turn. Although there is a partial train/test mismatch, this observation suggests that the older history turns are largely irrelevant to generating the response, and their presence does not significantly distract the models. However, we report a significant loss in performance across metrics in the extreme case where no history is provided to the model during training (train: h=$\emptyset$). Through manual inspection of samples, we hypothesize that this is due to the model treating the entire history as a knowledge snippet and attempting to ground the response under this assumption.

In evaluation configurations with $h\in\{3,2,1\}$, we observe that the performances of the original model and its respective variants are roughly similar, exhibiting only a slight decline as more history turns are removed. More precisely, ground truth response similarity metrics show a steady decline, while those measuring similarity with the provided knowledge slightly improve as fewer history turns are seen during training and/or evaluation. These results indicate that the conversation history is ignored not only during inference but also during model training. 

However, when the entire conversational history is omitted during evaluation (eval: h=$\emptyset$), the original model produces responses that are slightly better aligned with the ground truth response, showing a +$0.4$ gain in BLEU score and a +$0.5$ gain in BERTScore compared to the model trained without conversation history (train: h=$\emptyset$). Notably, both models achieve the best alignment with the given knowledge results so far, reaching $82.1$\% and $83.9$\% respectively, and report a $0$\% hallucination ratio as per the FaithDial \critic{}. It is worth mentioning that the model trained without any history performs significantly better across all four metrics when the history is also discarded during inference. This suggests that this model is most likely learning primarily to paraphrase a knowledge snippet into its response. From these analyses, we conclude that both the annotation strategies and evaluation methods of FaithDial do not take into account a crucial scenario in information-seeking dialogue, where a valid response depends on understanding and reasoning about the conversation history.

\section{\testbed{}}
\label{sec:charp}
\testbed{}, the proposed diagnostic set, exclusively assess whether information-seeking dialogue systems effectively attend to and use the conversation history. \testbed{} is built by modifying examples from the FaithDial validation set to ensure maximum domain alignment with FaithDial and to minimize annotation costs. That is, we edit FaithDial examples to make their response dependent on the conversation history analogously to FaithDial's editing of WoW annotations to make them hallucination-free. It is important to note that the FaithDial validation and test sets are sampled from the same distribution and exhibit similar result patterns. We create two variants of \testbed{}: \htestbed{} (\S~\ref{sec:hCHARP Creation}) for examples where addressing the last seeker's inquiry requires reasoning over the conversation history, and \etestbed{} (\S~\ref{sec:eCHARP Creation}), where the last inquiry can be addressed without such reasoning. We annotate 42\% of the FaithDial validation set (after excluding examples without conversation history) 
containing $2,160$ examples, split equally between \htestbed{} and \etestbed{}.

\subsection{\htestbed{} Creation}
\label{sec:hCHARP Creation}

In \htestbed{}, which refers to \emph{hard} \testbed{}, examples are expected to test basic natural language understanding abilities, with the expectation that the response will be straightforward if a model is attentive to the conversation history. An example is designed to test the ability to resolve co-reference relations and the history mentions \quotes{\textit{my favorite color is red}}, then the last user turn might be \quotes{\textit{Wondering which fruit has the same color as my favorite?}}. In other examples, annotators may introduce temporal reasoning (e.g., \textit{the pre-historic era is before the 16th century}), geospatial (e.g., \textit{Paris} is located in \textit{France}), or taxonomic (e.g., \textit{pie} is a type of \textit{patisserie}). To ensure systematic and high-quality annotations, we define a set of edit rules for each part of the example (ref \S~\ref{sec:Annotation Rules}).

\subsection{\etestbed{} Creation}
\label{sec:eCHARP Creation}

We create a set of domain control examples ---\etestbed{} for \emph{easy} \testbed{}--- that contains the same examples in \htestbed{}, but with the last user turn rewritten to be self-contained and independent from the conversation history. For instance, if the knowledge lists fruits that are typically green and others that are red, the last user turn would be \quotes{\textit{Wondering which fruit is typically red?}}. Thus, responding to examples in \etestbed{} should be easy. Poor performance on such examples\footnote{e.g., a system that copies or rephrases the entire knowledge and ignores even the last user turn.} suggests a domain (or task definition) shift between \testbed{} and the information-seeking dataset on which the model is trained. 
Further, in both the versions we annotate the \emph{knowledge} in the FaithDial dataset to provide a relevant, and a distracting factual information. The distracting information is designed to be ignored if the conversation history is considered in the knowledge selection by the models.

\section{Results}

\subsection{\testbed{} Automatic Evaluation}
\label{sec:charp-auto-eval}

The performances of the \flan{-base} and \godel{-base} models on \testbed{}, and on the subset of FaithDial validation examples that were utilized to construct \testbed{} are shown in \autoref{tab:auto_eval}. First, we observe that the distribution of performances on the validation set are similar to the test set (\autoref{tab:baseline_faithdial}) as they are sampled from the same distribution. Further, the results on the full validation set align with those on the subset used to build \testbed{} indicating that the sampled set from FaithDial used in \testbed{} does not introduce any bias to this study. 

\begin{table}[!htp]
    \begin{center}
    
    \resizebox{\columnwidth}{!}{
\begin{tabular}{lcccc}
\toprule
\multirow{2}{*}{\bf Models} & {\footnotesize \textbf{BLEU}$\uparrow$} & \footnotesize{\textbf{Critic $\downarrow$}} & \multicolumn{2}{c}{\footnotesize \textbf{BERTScore}$\uparrow$} \\
& {{ $(y, y')$}} & {{ $(k, y')$}} & {{ $(y, y')$}} & {{ $(k, y')$}}\\
\midrule

\multicolumn{5}{c}{\textit{FaithDial Valid.}} \\
\midrule
\flan{-base} & $14.6$      & 0.3           & 70.6       & 80.7        \\
\godel{-base}  & 14.8      & 0.3           & 70.8       & 80.6        \\
\midrule

\multicolumn{5}{c}{\textit{FaithDial Valid. (\testbed{} subset)}} \\
\midrule
\flan{-base} & 14.6      & 0.4           & 71.1       & 81.2        \\
\godel{-base} & 14.5      & 0.3           & 70.8       & 81.5 \\

\midrule
\multicolumn{5}{c}{\textit{\etestbed{}}} \\
\midrule
\flan{-base} & 22.8      & 1.7           & 70.8       &  79.4        \\
\godel{-base} & 20.5      & 0.5           & 69.4       &  82.5   \\

\midrule
\multicolumn{5}{c}{\textit{\htestbed{}}} \\
\midrule
\flan{-base} & 22.0      &   1.9         & 70.1       &   78.6  \\
\godel{-base} & 18.7      & 0.6           & 67.6       & 81.7    \\

\bottomrule
\end{tabular}

    }	
 \end{center}	
	
\caption{Performance of models on the FaithDial dataset across four evaluation sets: FaithDial validation set, a subset of the FaithDial validation set used to build \testbed{}, \etestbed{}, and \htestbed{}.  All scores are scaled within the range of $[0, 100]$.}
\label{tab:auto_eval}
\end{table}

Unsurprisingly, we observe that the models perform well on \etestbed{} with better BLEU scores in $(y,y')$, and almost similar on both {{ $(k, y')$}} metrics compared to the results on the validation set. The high BLEU scores arise because our ground truth responses have a high lexical overlap with the knowledge, involving less paraphrasing, compared to the FaithDial data. As the with and without paraphrasing the responses are semantically similar, the BERTScores remain similar to the one on the validation set. We also notice that \godel{-base} is less hallucinatory than \flan{-base}, performing better on both {{ $(k, y')$}} metrics. Conversely, \flan{-base} outperforms \godel{-base} on both {{ $(y, y')$}} metrics. 

Despite the \htestbed{} responses being strongly dependent on information from the conversational history, we notice that the models perform surprisingly well. When comparing the score ranges to those on \etestbed{}, we observe that, although the results are consistently lower across all metrics the difference was not significant. These observations strongly contradict our hypothesis in \S~\ref{sec:Probing History Awareness}, which posits that models ignoring conversational history should incur significant penalties across all metrics. We examine the metrics themselves by computing the \critic{} and \bert{Score} between the knowledge snippets and the gold responses in both the FaithDial validation and test sets, as well as in \testbed{}. 

\begin{table}[!htp]
    \begin{center}
    
    \resizebox{\columnwidth}{!}{
\begin{tabular}{lccc}
\toprule
& \bf Valid & \bf Test &  \bf \testbed{} \\
\midrule

\footnotesize \textbf{Critic} {{ $(k, y)\downarrow$}} & 0.4 & 0.4 & 16.0 \\
\footnotesize \textbf{BERTScore}{{ $(k, y)\uparrow$}} & 84.3 & 85.6 & 69.9 \\

\bottomrule
\end{tabular}

    }	
 \end{center}	
	
\caption{Evaluation of the ground truth response ($y$) when contrasted with the knowledge snippet ($k$) on FaithDial validation and test sets as well as on \testbed{}.  All scores are scaled within the range of $[0, 100]$.}
\label{tab:auto_eval_gold}
\end{table}

Results in \autoref{tab:auto_eval_gold} show that the ground truth responses of \testbed{} are labeled as extremely hallucinatory beyond not only the other gold responses in the compared sets of FaithDial but also to model predictions. The hallucination ratio is $27\times$ higher for the ground truth ($16$\%) compared to the response generated by \godel{-base} ($0.6$\%). Additionally, the semantic similarity with the knowledge, as measured by the \bert{}Score, is roughly more than 12\% lower between the ground truth ($69.9$\%) and \godel{-base} response ($81.7$\% on \htestbed{}). This contrasts with the scores on the FaithDial evaluation sets, where we observe a close tie with the models' responses. These observations indicate possible deficiencies in the metrics used in FaithDial and their comprehensiveness as the success through the lens of the metric lies at models focusing on the knowledge segment.

\subsection{\testbed{} Human Evaluation}
\label{sec:Human Evaluation}

We employ our human annotators (ref \S \ref{sec:app:Annotation Guideline} for details) to carry out a comprehensive analysis of the models' outputs. We focus on errors related to the system's reasoning over knowledge and conversation history. Through this evaluation, we emphasize faithfulness to the provided knowledge while including aspects such as cooperativeness, engagingness, and abstractiveness as motivated in \cite{dziri2022faithdial}. We frame the evaluation process in the form of a checklist ticker (binary classification), where each annotator is tasked with labeling whether a system response: 

\begin{itemize}
    \item $\mathbfcal{C}1$ properly addresses the seeker comment.
    \item $\mathbfcal{W}1$ addresses the seeker's comment while adding extra information not in the provided knowledge.
    \item $\mathbfcal{W}2$ is simply a copy or slight paraphrasing of the entire knowledge, while part of it is irrelevant.
    \item $\mathbfcal{W}3$ states a lack of knowledge (e.g. \emph{I don't know}) but still copies or rephrases part or all of the provided information, despite the relevant information existing within the provided knowledge.
    \item $\mathbfcal{W}4$ is a copy or slight paraphrasing of an irrelevant knowledge segment.
    \item $\mathbfcal{W}5$ fuses knowledge segments leading to wrong or contradictory information.
    \item $\mathbfcal{W}6$ is incorrect for other reasons, e.g. fully detached, severe hallucination, contains contradictory information.
\end{itemize}
    
\autoref{tab:human_eval_testbed_base} shows the human evaluation results of FaithDial trained \flan{-base} and \godel{-base} models on \testbed{} subsets. First, we observe that both models exhibit poor performance on \testbed{}, which slipped through FaithDial's automatic metrics (\autoref{tab:auto_eval}). 
As expected, \htestbed{} proves to be more challenging than \etestbed{}, with the correct response ratio ($\mathcal{C}1$) significantly dropping by $13$\% for \flan{-base} and $8$\% for \godel{-base}.

\begin{table}[!htp]
    \begin{center}
    
    \resizebox{\columnwidth}{!}{
\begin{tabular}{l|cc|cc}
\toprule
& \multicolumn{2}{c|}{\textbf{\flan{-base}}} & \multicolumn{2}{c}{\textbf{\godel{-base}}} \\
& \footnotesize \etestbed{} & \footnotesize \htestbed{} & \footnotesize  \etestbed{} & \footnotesize \htestbed{} \\
\midrule
$\mathcal{C}1$\hspace{2mm}{{ $\uparrow$}}  &  23\%         & 10\%         & 21\%                     & 13\%                     \\
\hdashline
$\mathcal{W}1${{ $\downarrow$}}  & 0\%          & 0\%          & 0\%                      & 1\%                      \\
$\mathcal{W}2${{ $\downarrow$}}  & 49\%         & 49\%         & 34\%                     & 36\%                     \\
$\mathcal{W}3${{ $\downarrow$}}  & 3\%          & 14\%          & 7\%                      & 12\%                     \\
$\mathcal{W}4${{ $\downarrow$}}  & 20\%         & 20\%         & 32\%                     & 31\%                     \\
$\mathcal{W}5${{ $\downarrow$}}  & 4\%          & 7\%          & 6\%                      & 7\%                      \\
$\mathcal{W}6${{ $\downarrow$}}  & 1\%          & 0\%          & 0\%                      & 0\%  \\
\bottomrule
\end{tabular}

    }	
 \end{center}	
	
\caption{Human evaluation results of \godel{-base} and \flan{-base} models on \etestbed{} and \htestbed{}.}
\label{tab:human_eval_testbed_base}
\end{table}

\begin{table*}[!htp]
    \begin{center}
    
    \resizebox{\textwidth}{!}{
\begin{tabular}{l|cc|cc|cc|cc|}
\cline{2-9}
& \multicolumn{2}{c|}{\textit{Finetuning}} & 
\multicolumn{6}{c|}{\textit{3-shot}} \\ \cline{2-9}
& \multicolumn{2}{c|}{\textbf{\llama{}}}  & 
\multicolumn{2}{c|}{\textbf{\llama{}}} & 
\multicolumn{2}{c|}{\textbf{\mixtral{}}} & 
\multicolumn{2}{c|}{\textbf{\chatgpt{}}}\\

& \footnotesize \etestbed{} & \footnotesize \htestbed{} & 
\footnotesize  \etestbed{} & \footnotesize \htestbed{} & 
\footnotesize  \etestbed{} & \footnotesize \htestbed{} & 
\footnotesize  \etestbed{} & \footnotesize \htestbed{} \\
\midrule

$\mathcal{C}1$\hspace{2mm}{{ $\uparrow$}}    &  36\%               & 27\%                                        & 26\%            & 13\%            & 71\%                       & 64\%                       & 66\% & 56\% \\
\hdashline
$\mathcal{W}1${{ $\downarrow$}}  & 0\%                & 0\%                                           & 34\%            & 37\%            & 18\%                       & 14\%                        & 18\% & 13\% \\
$\mathcal{W}2${{ $\downarrow$}}  & 32\%               & 35\%                                          & 12\%            & 9\%            & 4\%                        & 7\%                       & 3\%  & 6\% \\
$\mathcal{W}3${{ $\downarrow$}}  & 7\%                & 13\%                                         & 0\%             & 0\%             & 0\%                        & 0\%                        & 2\%  & 3\%\\
$\mathcal{W}4${{ $\downarrow$}}  & 20\%               & 19\%                                         & 0\%             & 0\%             & 1\%                        & 2\%                        & 0\%  & 1\%\\
$\mathcal{W}5${{ $\downarrow$}}  & 4\%                & 5\%                                          & 7\%             & 9\%             & 4\%                        & 9\%                        & 5\%  & 9\% \\
$\mathcal{W}6${{ $\downarrow$}}  & 1\%                & 1\%                                           & 21\%            & 32\%            & 2\%                        & 4\%                        & 6\%  & 12\% \\
\bottomrule
\end{tabular}

    }	
 \end{center}	
	
\caption{Human evaluation results on \testbed{} for models under fine-tuning, 3-shot learning paradigms.}
\label{tab:human_eval_testbed_few_shot}

\end{table*}

We see that models do not add out-of-context information ($\mathcal{W}1$) or suffer from severe hallucinations ($\mathcal{W}6$), as these error rates are almost null across all configurations. This is largely expected as FaithDial training reinforces this in models. We also note that $60$\% the responses contain paraphrased \emph{all} knowledge including the irrelevant fact $\mathcal{W}2$, or only the irrelevant knowledge $\mathcal{W}4$. While these samples are marked as errors by humans, automatic metrics struggle to identify the same (\autoref{tab:auto_eval_gold}).
As spotting the knowledge chosen is relevant or not is contingent on knowing the conversation so far, metrics focusing on only the knowledge fails unlike humans considering the context as well.

We observe that only $\mathcal{W}3$ errors show a significant increase when comparing the performances on \etestbed{}  and \htestbed{}: $11$\% for \flan{-base} and $5$\% for \godel{-base}, respectively. This observation is particularly interesting as it directly relates to the FaithDial annotation guide, which instructs the annotators to write responses where the bot acknowledges its ignorance and continues the conversation by presenting the given knowledge engagingly when the knowledge cannot satisfactorily address the seeker's last inquiry. This means that a model that finds a knowledge to be relevant for an example in \etestbed{}, may find it irrelevant in its corresponding example in \htestbed{} that is attributed to FaithDial trained models' shortcoming to reason over the conversation history.  

Interestingly, we notice that the performances are roughly equal under some categories (mainly $\mathcal{W}2$ and $\mathcal{W}4$) when comparing \etestbed{} and \htestbed{} results. This is noteworthy because, in \etestbed{}, models do not need to rely on previous conversation history to respond, as the last utterance is designed to be self-contained. In contrast, \htestbed{} is designed to assess whether models consider the entire conversation history. For example, a model that simply copies the entire knowledge segment ($\mathcal{W}2$), without considering the content of the last user utterance, is effectively ignoring the entire history. This observation suggests that the errors noted are not due to the model’s inability to reason based on earlier conversation turns.

\section{Analysis}

\subsection{\textit{On FaithDial Data Artifact}}

We conduct ablations on the behavior of models to estimate the resulting errors due to the artifacts in the FaithDial training data by comparing the performances of models with and without fine-tuning on FaithDial. We use the few-shot learning technique via prompting to ablate for not training with FaithDial dataset.
Specifically, we compare the performances of \llama{} model that we tuned on FaithDial against the same model with $3$ in-context examples (\textit{3-shot}). In addition, we evaluate the \textit{3-shot} performances of \chatgpt{}~\cite{chatgpt2022} and \mixtral{}~\cite{jiang2024mixtral} LLMs. Implementation details of these experiments can be found in Appendix~\ref{sec:Few-shot Experiments}, as well as an example of the models' responses in \autoref{fig:eval_example}.   

In \autoref{tab:human_eval_testbed_few_shot} we compare the human evaluation results on \etestbed{} and \htestbed{}\footnote{We also measured inter-annotator agreements across all models and sets, and have reported the results in \S ~\ref{sec:Inter-annotator Agreement}.} across the different error types. While the fine-tuned \llama{} reports higher
\footnote{although it underperforms on automatic evaluation metrics. This aspect is further discussed in Appendix~\ref{sec:Automatic Evaluation}.} performance ($\mathcal{C}1$) than \flan{-base} and \godel{-base} (\autoref{tab:human_eval_testbed_base}), due to its larger size, we believe it also suffers from a lack of reasoning behavior. This is suggested by the reported error trends, where $\mathcal{W}2$ and $\mathcal{W}4$ are the dominant error categories, in contrast to $\mathcal{W}1$, $\mathcal{W}5$, and $\mathcal{W}6$. However, the error trends undergo a drastic change when comparing the results of fine-tuned \textit{3-shot} models. 

First, we observe that $\mathcal{W}1$ is the dominant error category, where models add extra information after addressing the user inquiry. We attribute this behavior to the verbose nature of LLMs, which is challenging to mitigate without further tuning~\cite{gudibande2023false}. Second, we notice that the error ratio for $\mathcal{W}2$ is significantly lower, by at least $20$\%, across configurations compared to fine-tuned models. Additionally, we observe near-zero error ratios for $\mathcal{W}3$ and $\mathcal{W}4$, strongly suggesting that models not tuned on FaithDial are not affected by their ability to reason over the history. The error trend of \llama{} trained on FaithDial strongly mimicking the results in \autoref{tab:human_eval_testbed_base}, we confirm with a high probability that the LMs lose the ability to account for conversation after being fine-tuned on FaithDial.

Third, non-tuned models suffer significantly more from severe hallucinations ($\mathcal{W}6$), a well-known issue with LLMs~\cite{ji2023survey,ye2023cognitive}. However, this issue tends to be mitigated as the models get larger; for instance, $\mathcal{W}6$ drops from $32$\% in \llama{} to $5$\% in \mixtral{} on \htestbed{}. While the smaller \llama{} LLM performs worse than its fine-tuned version, the larger models, \mixtral{} and \chatgpt{}, significantly outperform all reported fine-tuned models by at least $30$\% on $\mathcal{C}1$. Despite their high performances, we observe that \htestbed{} remains more challenging than \etestbed{} for even the most advanced models, indicating that \testbed{} can additionally serve as a measure of reasoning capability for such models. Finally, the fact that \mixtral{} outperforms \chatgpt{} in our tasks serves as an additional indicator that high performance is achievable through community-shared, open-source models.

\subsection{\textit{On FaithDial Evaluation Metrics}}

Despite being highly accurate, human evaluation is time and resource-consuming which limits its scalability and practicality on large evaluation sets. To this end, we investigate the recent trend~\cite{wang2023chatgpt,liu2023gpteval,hackl2023gpt,li2024leveraging} of utilizing LLM APIs for the open-ended evaluation of NLP systems. More specifically, our focus is on using \gptfturbo{}, a cost-efficient version of GPT-$4$~\cite{OpenAI2023GPT4TR}. This family of models has been shown to correlate with human judgment, outperforming other alternatives~\cite{chiang2023vicuna,liu2023g} in their generation. The exact prompt we used and a detailed description of this experiment are presented in Appendix~\ref{sec:GPT-4 Evaluation}.

In \autoref{fig:heatmap_main} the normalized contingency table is displayed as a heatmap showcasing the agreement between \gptfturbo{} and human judgments of the \llama{} fine-tuned models on both \etestbed{} and \htestbed{}. The table counts the frequency of each combination of categories from \gptfturbo{} and human judgments, which we later normalized into percentages ($[0-100]$). We show the contingency table corresponding to fine-tuned \llama{} model, and other models\footnote{Detailed plots for all six models, as well as kappa agreement scores, can be found in Figure~\ref{fig:heatmap_app} in the Appendix~\ref{sec:GPT-4 Evaluation}.} also show a similar trend. 

\begin{figure}[!ht]
    \centering
    \includegraphics[scale=0.23]{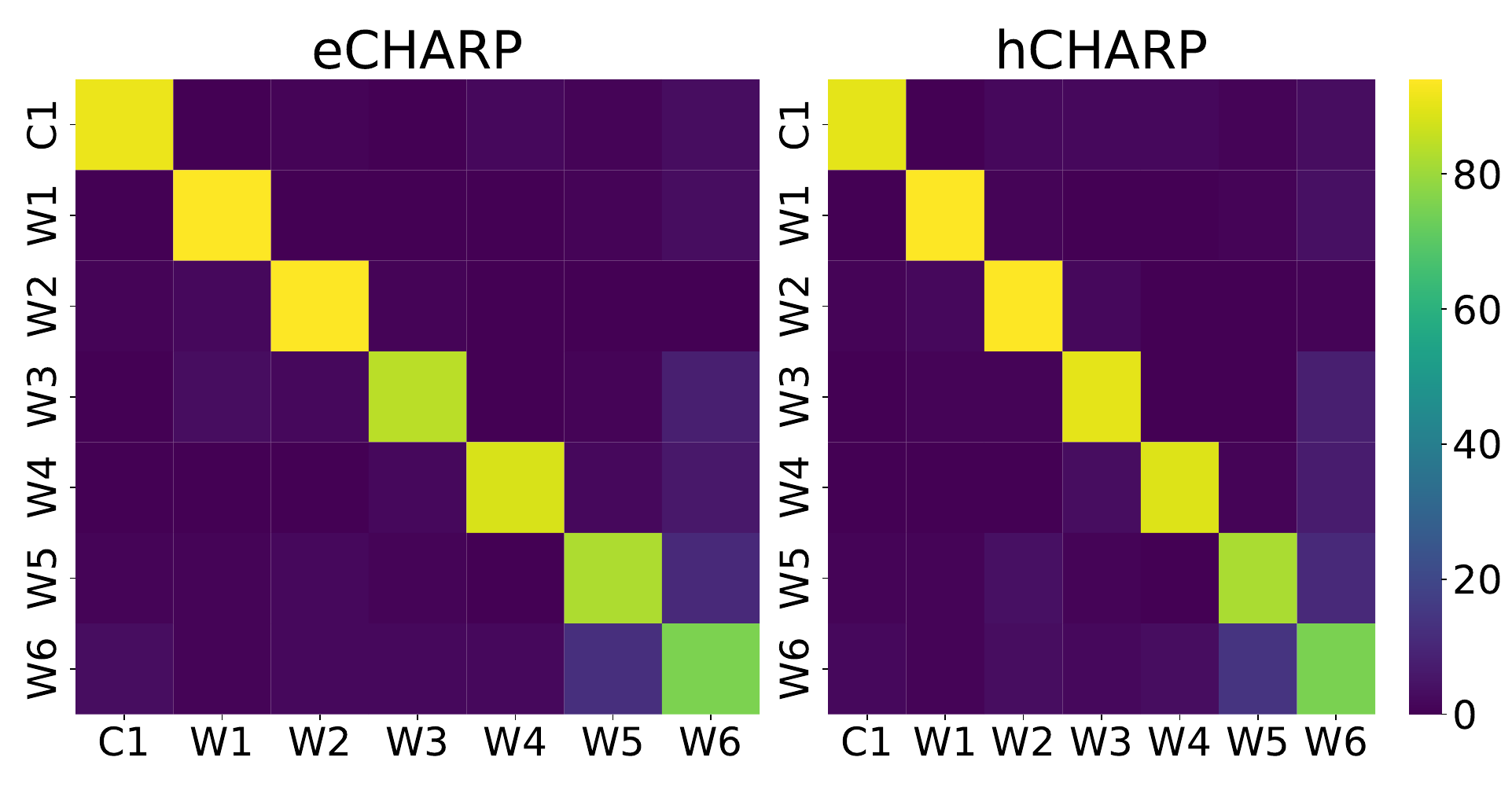}
    \caption{Heatmap showing the normalized (percentage) contingency tables of evaluation categories between \gptfturbo{} (rows) and human (columns) judgments. It was measured on the output of \llama{} (finetuned) for both \etestbed{} (left) and \htestbed{} (right).}
    \label{fig:heatmap_main}
\end{figure}

First, it is worth mentioning that the Kappa~\cite{carletta1996assessing} agreement score on the overall examples of \etestbed{} and \htestbed{} are $0.89$ and $0.88$, respectively. These scores are higher than the $0.8$ well-acceptable threshold, indicating a high overall correlation. This observation is further evidenced by the high values along the diagonals of the correlation heatmaps for both subsets. More precisely, we observe that the correlation is higher for the correct response category (\textbf{$\mathcal{C}1$}), as well as for other wrong response categories that are relatively easy to detect, such as $\mathcal{W}1$, $\mathcal{W}2$, and $\mathcal{W}4$. However, we observed that \gptfturbo{} tends to confuse certain categories, notably $\mathcal{W}3$ (which involves stating lack of knowledge while still providing relevant information) and $\mathcal{W}5$ (fusing the irrelevant knowledge segments), with the severe hallucination category ($\mathcal{W}6$). Although not a perfect match, we believe that powerful LLM currently represents the best approximation to human annotations instead of the weak automatic evaluation metrics. 

\section{Conclusion}

In this work, we examine the impact of annotation artifacts on information-seeking dialogue models tuned on FaithDial, a well-established, hallucination-free annotation benchmark. We introduce \testbed{}, a diagnostic set designed to evaluate the ability of models to reason over the conversation history, while also staying grounded on the knowledge. Our analysis with \testbed{} reveals a strong correlation between training on FaithDial to models' ignoring reasoning over the conversation history. Further, proprietary LLM APIs can be a proxy to human evaluation, and a better hallucination estimator to automatic metrics. In similar vein to \cite{chen2023chatgpt}, we note that while it is important to ensure hallucination-free annotations, including examples to cover reasoning over context and other pretraining knowledge is necessary to preserve models' reasoning capabilities.

\section*{Limitations}

Potential limitations of this work could be stemming from the sampling of dataset to conduct the study. Although the study focuses primarily on FaithDial dataset, the other existing datasets have been shown to contain more hallucinations rendering this a minor issue. Further, the knowledge grounded dialogue generation has not looked at the generated texts that pertains to a diverse demographics. This is largely due to the nacency of this domain and further studies can alleviate this issue.

\section*{Acknowledgements}
We would like to thank Imad Mousaoui, Ella Cho, Abdulmuizz Yusuf, and Parminder Singh Bharot, the professional annotators without whom this work would have not been possible. We thank the anonymous reviewers for their insightful comments.

\bibliography{custom}


\appendix

\begin{figure*}[ht]
    \centering
    \includegraphics[width=\textwidth]{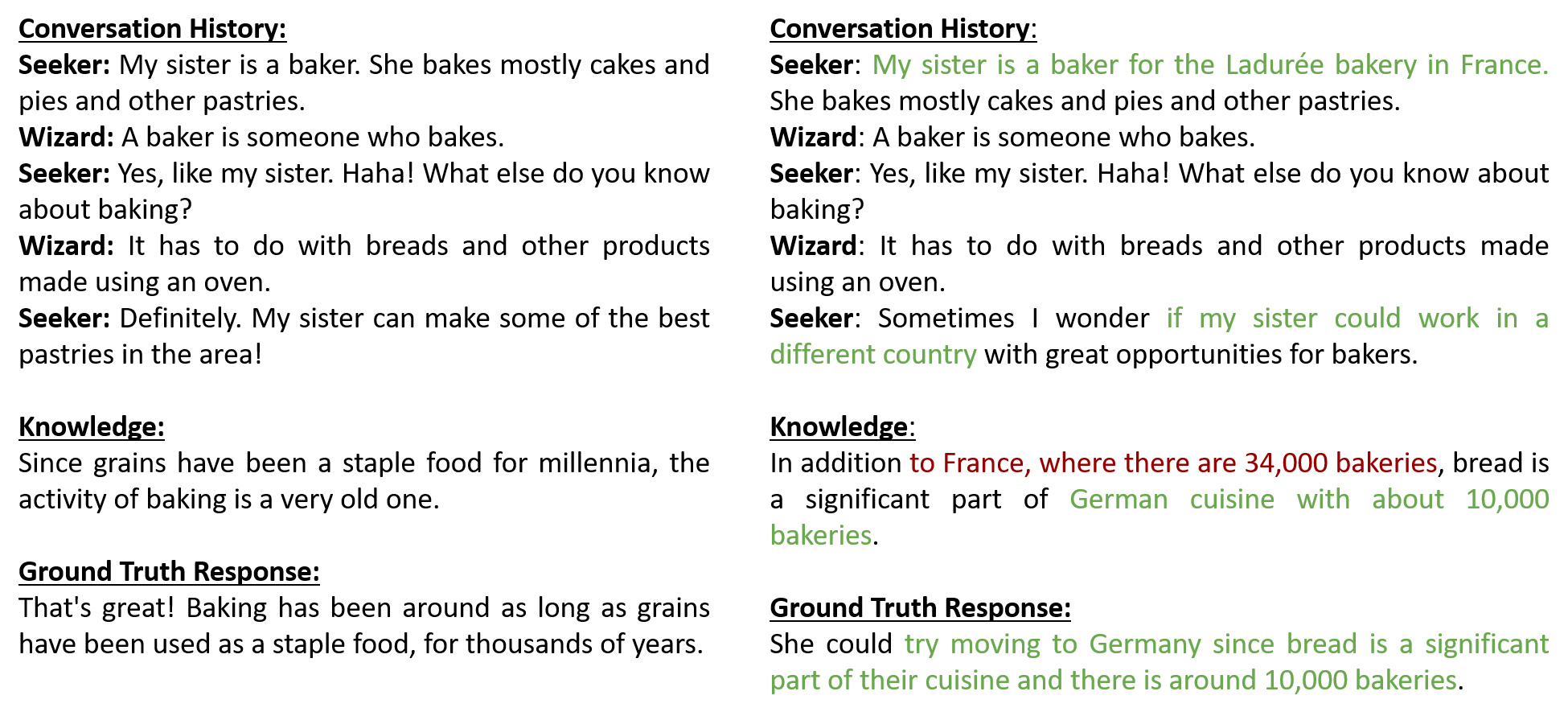}
    \caption{Original example from the FaithDial validation set (left) and our edited \htestbed{} version (right). Green text indicates content that the model is expected to reason over, while red text marks distracting content within the provided knowledge.}
    \label{fig:anno_pair}
\end{figure*}

\section{Annotation Guideline}
\label{sec:app:Annotation Guideline}

Although editing is generally faster than creating new examples from scratch, the numerous constraints that must be met in a single example could make the task time-consuming if annotators are not provided with \textit{template-like} instructions. Therefore, we decided to restrain our annotators to introducing edits that probe two natural language understanding abilities: solving co-reference relations and performing simple reasoning. Figure~\ref{fig:anno_pair} presents an example from the FaithDial validation set, transformed into an \htestbed{} annotation according to the rules below.

\subsection{Annotation Process}

We hire $4$/ $20$ interviewed annotators as contractors over prior experience and a predefined test. For the $\approx 5700$ annotation hours, annotators were paid $19$ USD/hour. The annotators were trained beforehand and given guidelines containing instructions and examples of both typical and radical cases they might encounter during annotation. In addition, domain experts were revising the annotations daily and conducting video meetings with annotators whenever necessary. Typically, an annotator would receive an example comprising $<$conversation history; last user turn; knowledge; response$>$, which he/she is required to edit according to the guidelines outlined in the following section. Despite our efforts to simplify the annotation task and guide the annotators, the process was still considered slow, with annotators averaging only 8 examples per hour. This slow pace was due to the need to adhere to both the original FaithDial guidelines and all our additional conditions outlined below.

\subsection{Annotation Rules}
\label{sec:Annotation Rules}

\noindent \textbf{Conversation history} We give the annotators the freedom to fully rewrite the response, knowledge, and last user turn, while not introducing changes to the conversation history unless they find it necessary (and if so, to make the minimal possible edits). Also, we strongly encourage them to maintain the natural flow of the conversation and stick to the information-seeking dialogue style.

\noindent \textbf{Last user turn} The last user turn should straightforwardly seek a specific piece of information and be answerable only when referencing the conversation history. We instruct our annotators to avoid user requests that may elicit multiple valid responses with different semantic meanings, as these are not easily measurable with automatic metrics.

\noindent \textbf{Knowledge} The knowledge should maintain the same properties as the original, providing correct factual information (directly relevant to user request) in 1-2 sentences with a maximum of 30 words. We found it practical to structure the knowledge with two pieces of information: one distractive and the other relevant to the last user turn. The distractive element should be easy to ignore in the response if the model adequately attends to the conversation history.

\noindent \textbf{Response} The response to the last user turn should be the only unique and valid response, based on the information contained in the provided knowledge. Additionally, in line with FaithDial guidelines, the response should be faithful to this knowledge, often comprising a large portion that is either a direct copy or a paraphrase of it. However, we instructed our annotators to perform only minimal paraphrasing necessary to ensure a well-structured response. We do so to save annotation time, as knowledge rephrasing isn't the objective of \testbed{}. Additionally, this helps avoid evaluation mismatches caused by incidental inconsistencies between the annotation style of our annotators and that of the FaithDial crowd-workers.

\section{Experimental Setting}
\label{sec:Experimental Setting}

\subsection{FaithDail \critic{}}
\label{sec:app:FaithDail Critic}

FaithDial~\cite{dziri2022faithdial} \critic{} was trained using a dataset comprising 14,000 hallucinatory turns (edited original WoW turns) and 20,000 faithful turns (unedited WoW and FaithDial turns), serving as negative and positive examples, respectively. More precisely, the authors paired up turns with their respective knowledge snippet and trained the RoBERTa-large model~\cite{liu2019roberta} by framing the task as sequence pair binary classification. The FaithDial hallucination detector not only demonstrates a high correlation with human judgment but also excels in hallucination detection testbeds like BEGIN~\cite{dziri2022evaluating}. Furthermore, it outperforms classifiers trained on counterpart hallucination detection datasets, such as DECODE~\cite{welleck2019dialogue} and DNLI~\cite{nie2021like}.

\subsection{Llama Finetuning}
\label{sec:Llama Finetuning}

\textbf{\llama{}} We utilize the 7B chat version of the Llama-2 model series~\cite{touvron2023llama}, which is the largest model we could effectively fine-tune (compared to the 13B and 70B versions), given our computational resources. We fine-tuned the model on a single node equipped with 8 NVIDIA V100 GPUs with 32GB of memory, utilizing a codebase built on the PyTorch version~\cite{paszke2019pytorch} of the Transformers library~\cite{wolf2020transformers}. The initial learning rate was set to 2e-6, employing the AdamW optimizer~\cite{kingma2014adam} with a cosine decay learning rate schedule. The model was trained over 5 epochs with a maximum sequence length of 1024 tokens. We set the per-GPU batch size to 48, the maximum size that we can fit on a single GPU. Training acceleration was achieved by leveraging the deepspeed library~\cite{rasley2020deepspeed}, mixed precision training~\cite{fp16}, and gradient checkpointing~\cite{chen2016training}. We pick up the best checkpoint using early stopping based on performance on FaithDial validation set.

\subsection{Few-shot Experiments}
\label{sec:Few-shot Experiments}

In addition to \llama{}, we also conduct out-of-the-box inference (without fine-tuning) on experiments using gpt-3.5-turbo~\cite{chatgpt2022} and Mixtral-8x7B~\cite{jiang2024mixtral}. Throughout this paper, we refer to these models as \chatgpt{} and \mixtral{}, respectively. To this end, we carefully 
design a prompt that takes the conversation history and the knowledge relevant to the last seeker's turn as input to generate a bot response:

{\small

\texttt{You are given a chitchat conversation between a ``User'' and a ``Bot''. Your goal is to generate a response to the last user turn, which in turn should be based on the given ``Knowledge''. You are prohibited from generating any extra information that is not mentioned in the given knowledge. The output should be a JSON dictionary as follow: \{``response'': ``''\}. Here are a few demonstration examples:}

\vspace{1em}
\texttt{[In\_CONTEXT\_EXAMPLE\_1]}

\vspace{1em}
\texttt{[In\_CONTEXT\_EXAMPLE\_2]}

\vspace{1em}
\texttt{[In\_CONTEXT\_EXAMPLE\_3]}

\vspace{1em}
\texttt{[INPUT\_EXAMPLE]}

\normalsize

We designed the instruction part of the prompt through trial and error iterations until we verified that all models could follow the instructions and generate a response that addresses the user query in our required format. Then, we continuously added in-context examples until the output of all models stabilized (with minor to no changes in the model response). We set the number of in-context examples, that were picked up from FaithDial training set, to 3 as we didn't see any improvement in adding more examples or performing more prompt engineering. On one hand, we execute the generation process of \chatgpt{} and \mixtral{} samples through the commercial APIs of OpenAI~\footnote{https://chat.openai.com/} and Replicate~\footnote{https://replicate.com/}, respectively. On the other hand, we use our local V100 GPUs to infer with \llama{}. However, across all experiments, we set the temperature to 1.0, the frequency penalty to zero, and top-p to 1.0, aiming to minimize randomness during the generation process.

\section{Analysis}

\subsection{Automatic Evaluation Results}
\label{sec:Automatic Evaluation}

Tables \ref{tab:auto_eval_valid}, \ref{tab:auto_eval_echarp}, and \ref{tab:auto_eval_hcharp} show the automatic metric scores of the models fully tuned on FaithDial and under the \textit{3-shot} setting on the FaithDial validation subset, \etestbed{}, and \htestbed{}, respectively. First, we observe that the finetuned \llama{}, across all three evaluation sets, systematically yields slightly worse results on all FaithDial metrics compared to \godel{-base} and \flan{-base}. We believe this is primarily because, despite full parameters tuning on FaithDial, \llama{} has retained some of its \textit{chatty behavior} that was induced during the SFT and RLFH training procedures. However, this does not mean that the outputs of \llama{} are of lower quality than those of \godel{-base} or \flan{-base}; in fact its the opposite as indicated by the human evaluation results in Tables \ref{tab:human_eval_testbed_base} and \ref{tab:human_eval_testbed_few_shot}. This particular observation aligns with the findings of other studies~\cite{sankar2019neural,yeh2021comprehensive,parthasarathi2020evaluate} regarding the limitations of automatic metrics in evaluating dialog systems.

\begin{table}[!htp]
    \begin{center}
    
    \resizebox{\columnwidth}{!}{
\begin{tabular}{lcccc}
\toprule
\multirow{2}{*}{\bf Models} & {\footnotesize \textbf{BLEU}$\uparrow$} & \footnotesize{\textbf{Critic $\downarrow$}} & \multicolumn{2}{c}{\footnotesize \textbf{BERTScore}$\uparrow$} \\
& {{ $(y, y')$}} & {{ $(k, y')$}} & {{ $(y, y')$}} & {{ $(k, y')$}}\\

\midrule
\multicolumn{5}{c}{\bf Finetuning} \\
\midrule

\flan{-base} & 14.6 & 0.4 & 71.1 & 81.2 \\
\godel{-base} & 14.5 & 0.3 & 70.8 & 81.5 \\
\llama{}   & 12.0 & 2.0 & 69.2 & 73.1 \\

\midrule
\multicolumn{5}{c}{\bf 3-shot} \\
\midrule

\llama{} & 3.7 & 72.9 & 54.3 & 59.5\\
\mixtral{} & 9.4 & 29.1 & 65.9 & 74.3 \\
\chatgpt{} & 6.5 & 55.2 & 62.3 & 67.6 \\


\bottomrule
\end{tabular}

    }	
 \end{center}	
	
\caption{Performance of models on FaithDial validation set used to build \testbed{}. full fine-tuning on FaithDial, and with no fine-tuning by using 3 \textit{in-context} examples. All scores are scaled within the range of $[0, 100]$. }
\label{tab:auto_eval_valid}
\end{table}

\begin{table}[!htp]
    \begin{center}
    
\resizebox{\columnwidth}{!}{
\begin{tabular}{lcccc}
\toprule
\multirow{2}{*}{\bf Models} & {\footnotesize \textbf{BLEU}$\uparrow$} & \footnotesize{\textbf{Critic $\downarrow$}} & \multicolumn{2}{c}{\footnotesize \textbf{BERTScore}$\uparrow$} \\
& {{ $(y, y')$}} & {{ $(k, y')$}} & {{ $(y, y')$}} & {{ $(k, y')$}}\\

\midrule
\multicolumn{5}{c}{\bf Finetuning} \\
\midrule

\flan{-base} & 22.0      &   1.9         & 70.1       &   78.6  \\
\godel{-base} & 18.7      & 0.6           & 67.6       & 81.7    \\
\llama{}    & 17.2      & 3.7           & 68.1       & 69.1        \\

\midrule
\multicolumn{5}{c}{\bf 3-shot} \\
\midrule

\llama{}  & 8.0       & 54.0          & 63.7       & 65.0        \\
\mixtral{} & 20.6      & 16.3          & 74.6       & 70.0        \\
\chatgpt{}  & 20.2      & 22.8          & 74.6       & 69.8  \\

                     
\bottomrule
\end{tabular}

    }	
 \end{center}	
	
\caption{Performance of models on \htestbed{}. All scores are scaled within the range of $[0, 100]$.}
\label{tab:auto_eval_echarp}
\end{table}

\begin{table}[!htp]
\begin{center}
    
\resizebox{\columnwidth}{!}{
\begin{tabular}{lcccc}
\toprule
\multirow{2}{*}{\bf Models} & {\footnotesize \textbf{BLEU}$\uparrow$} & \footnotesize{\textbf{Critic $\downarrow$}} & \multicolumn{2}{c}{\footnotesize \textbf{BERTScore}$\uparrow$} \\
& {{ $(y, y')$}} & {{ $(k, y')$}} & {{ $(y, y')$}} & {{ $(k, y')$}}\\

\midrule
\multicolumn{5}{c}{\bf Finetuning} \\
\midrule

\flan{-base} & 22.8      & 1.7           & 70.8       &  79.4        \\
\godel{-base} & 20.5      & 0.5           & 69.4       &  82.5   \\
\llama{}    & 20.1      & 3.9           & 70.0       & 68.8        \\

\midrule
\multicolumn{5}{c}{\bf 3-shot} \\
\midrule

\llama{} & 8.9       & 51.6          & 65.4       & 65.3        \\
\mixtral{} & 21.3      & 16.6          & 75.5       & 70.0        \\
\chatgpt{}  & 19.9      & 21.8          & 74.8       & 69.2 \\


\bottomrule
\end{tabular}

    }	
 \end{center}	
	
\caption{Performance of models on \etestbed{}. All scores are scaled within the range of $[0, 100]$.}
\label{tab:auto_eval_hcharp}
\end{table}

The results are much worse when comparing the \textit{3-shot} models with the fine-tuned ones across all metrics and evaluation sets. The high hallucination ratio, as indicated by the \critic{} score, is well-justified since these models (especially \llama{}) tend to incorporate out-of-knowledge information, a finding that is corroborated by human evaluation. However, our human evaluators noted that the responses from \mixtral{} and \chatgpt{} tend to be creative, often using different words than the provided knowledge. Despite this, they deliver responses that are semantically aligned with the given knowledge and have the same semantic meaning as the ground truth response. This tendency results in a misleadingly high hallucination ratio, suggesting that the FaithDial \critic{} model~\footnote{which was specifically-tuned on FaithDial examples, while \bert{Score} models, in contrast, were tuned on MNLI~\cite{williams2018broad}.} is overly sensitive to lexical overlapping and fails to capture the underlying semantic meaning. This is also noticeable when considering that \critic{} score increases more significantly than the drops in the \bert{Score} {{ $(k, y')$}}. For instance, while the \critic{} score increases by 70.9\%, the \bert{Score}{{$(k, y')$}} decreases by only 14.6\% when comparing the tuned \llama{} with its \textit{3-shot} counterpart on FaithDial validation subset. Still, FaithDial automatic metrics significantly under-estimate the performance of \mixtral{} and \chatgpt{} compared to fine-tuned \godel{-base} and \flan{-base}. However, it's interesting to note that the ranking of \textit{3-shot} models (\mixtral{} > \chatgpt{} > \llama{}) according to automatic metrics aligns with the ranking obtained through human evaluation.

\subsection{Inter-annotator Agreement}
\label{sec:Inter-annotator Agreement}

In an effort to assess the quality of human evaluations, we tasked our annotators to evaluate a subset of 64 randomly selected examples from both \etestbed{} and \htestbed{}. This evaluation covered all six model variants studied in our experiments, leading to 786 model outputs that were evaluated by 3 annotators. Table~\ref{tab:kappa_human_gpt_turbo} shows the inter-annotator agreement scores, as measured by the Kappa Coefficient~\cite{carletta1996assessing}.

\begin{table}[!htp]
    \begin{center}
    
\begin{tabular}{lcc}

\toprule
& \footnotesize \etestbed{} & \footnotesize \htestbed{} \\

\midrule
\multicolumn{3}{c}{\textit{Finetuning Models}} \\
\midrule
\flan{-base} & 93\% & 93\% \\
\godel{-base} & 93\% & 96\% \\
\llama{} & 94\% & 93\% \\

\midrule
\multicolumn{3}{c}{\textit{3-shot Models}} \\
\midrule

\llama{} & 88\% & 92\% \\
\mixtral{} & 97\% & 94\% \\
\chatgpt{} & 92\% & 89\% \\
\bottomrule

\end{tabular}

 \end{center}	
	
\caption{Kappa inter-annotator agreement scores reflect human judgments of three FaithDial-tuned models and three \textit{3-shot} models, based on a random set of 64 examples each from \etestbed{} and \htestbed{}.}
\label{tab:kappa_human_inter}

\end{table}

Overall, we observe significantly high agreement among annotators, well above the widely accepted threshold of 80\%. Despite slight variations, we notice that the kappa score remains above this threshold across all the configurations. This not only demonstrates the professionalism of our annotators but also the clarity and precision of our proposed evaluation schema.   

\subsection{GPT-4 Evaluation}
\label{sec:GPT-4 Evaluation}

Given the complete conversational history, knowledge, response, and a system's prediction, we constructed a prompt requiring \gptf{} to perform the same checklist evaluation procedure as outlined in~\S\ref{sec:Human Evaluation}:

{\small

\texttt{Your task is to assess the quality of a machine learning system's response in a conversation. The conversation 'history' includes interactions between a user (Seeker) and a bot (Wizard), along with relevant 'knowledge' that pertains to the Seeker's last utterance. You are also provided with a 'response' (a ground truth or expected response) and the 'prediction' (the system's predicted response). Your evaluation involves comparing the system's 'prediction' with the 'response', considering the entire conversation 'history' and the provided 'knowledge'.}

\texttt{The evaluation is structured around categorizing the system's response into specific categories. 
Your output should be a JSON dictionary with a single <key, value> pair. The key is "category", and the value is a list of the category numbers that the predicted response falls under. For example: {"category": [1]}. These categories are:}

\texttt{1. The system's prediction is of high quality and is an equivalent or a paraphrase of the ground truth response.}

\begin{center}
    \texttt{[In\_CONTEXT\_EXAMPLE\_1\_FOR\_CAT\_1]}
    \texttt{[In\_CONTEXT\_EXAMPLE\_2\_FOR\_CAT\_1]}
\end{center}

\texttt{2. The system's prediction aligns with the ground truth response, but adds extra information, meaning that the content in the prediction is absent from the ground truth response, the given knowledge or the given history.}

\begin{center}
    \texttt{[In\_CONTEXT\_EXAMPLE\_1\_FOR\_CAT\_2]}
    \texttt{[In\_CONTEXT\_EXAMPLE\_2\_FOR\_CAT\_2]}
\end{center}
\texttt{3. The system's prediction is an identical copy or a very similar rephrasing of the entire knowledge. Meaning that its content is the exact same that can be found in the knowledge. It might contain content that aligns with the ground truth response, but it also contains off-topic content from the knowledge. It should not contain information that is absent from the given knowledge.}

\begin{center}
    \texttt{[In\_CONTEXT\_EXAMPLE\_1\_FOR\_CAT\_3]}
    \texttt{[In\_CONTEXT\_EXAMPLE\_2\_FOR\_CAT\_3]} 
\end{center}

\texttt{4. The system's prediction states doubt and ignorance, saying it doesn't know, doesn't understand, or is not equiped to answer; yet it copies or rephrases the content from the knowledge. The system's prediction content may originate from the whole knowledge or from the part of the knowledge that correctly aligns with the ground truth response or from the part of the knowledge that is not aligned with the ground truth response. It should not contain information that is absent from the given knowledge. For example:}

\begin{center}
    \texttt{[In\_CONTEXT\_EXAMPLE\_1\_FOR\_CAT\_4]}
    \texttt{[In\_CONTEXT\_EXAMPLE\_2\_FOR\_CAT\_4]}  
\end{center}

\texttt{5. The system's prediction is an identical copy or a very similar rephrasing of the part of the knowledge whose content is off-topic and does not align with the ground truth response. The content of the system prediction should not contain any content that aligns (even partially) with the ground truth response. It should not contain information that is absent from the given knowledge or state ignorance by saying it doesn't know and is not able to get that information. }

\begin{center}
    \texttt{[In\_CONTEXT\_EXAMPLE\_1\_FOR\_CAT\_5]}
   \texttt{[In\_CONTEXT\_EXAMPLE\_2\_FOR\_CAT\_5]}
\end{center}

\texttt{6. The system's prediction does not align with the ground truth response and its content is made of mixed-up information coming from both the knowledge part that aligns with the ground truth response (on-topic) and the part that does not align with the ground truth response (off-topic). Both parts are not just a copy of the knowledge, but are merged together, which leads to wrong and inaccurate information in the prediction.}

\begin{center}
    \texttt{[In\_CONTEXT\_EXAMPLE\_1\_FOR\_CAT\_6]}
    \texttt{[In\_CONTEXT\_EXAMPLE\_2\_FOR\_CAT\_6]}
\end{center}
\texttt{7. The system's prediction does not align with the ground truth response yet it cannot be classified as any of the previously mentioned categories. This includes but is not limited to: having extra information that is absent from the given knowledge, having two or more content elements that contradict each other, being empty.}

\begin{center}
    \texttt{[In\_CONTEXT\_EXAMPLE\_1\_FOR\_CAT\_7]}
   \texttt{[In\_CONTEXT\_EXAMPLE\_2\_FOR\_CAT\_7]} 
\end{center}

\texttt{Now you must evaluate the following:}

\texttt{[INPUT\_EXAMPLE]}

}
\normalsize

We found that using two \textit{in-context} examples for each category works better than using just one, with no further improvement observed by using additional examples. Due to the high costs associated with calling \gptf{}, we opted for its more cost-effective version, \gptfturbo{}, to perform evaluations on the full evaluation sets.

\begin{table}[!htp]
    \begin{center}
    
\begin{tabular}{lcc}

\toprule
& \footnotesize \etestbed{} & \footnotesize  \htestbed{} \\

\midrule
\multicolumn{3}{c}{\textit{Finetuning Models}} \\
\midrule

\flan{-base} & 84\% & 84\% \\
\godel{-base} & 85\% & 87\% \\
\llama{} & 88\% & 89\% \\

\midrule
\multicolumn{3}{c}{\textit{3-shot Models}} \\
\midrule

\llama{} & 87\% & 86\% \\
\mixtral{} &  92\% &  90\% \\
\chatgpt{}  &  90\% &  91\% \\
\bottomrule

\end{tabular}

 \end{center}	
	
\caption{Kappa agreement scores between human judgments and those of \gptf{-turbo} regarding the quality of outputs from three FaithDial-tuned models and three \textit{3-shot} models on the full set of \etestbed{} and \htestbed{}.}
\label{tab:kappa_human_gpt_turbo}

\end{table}

Table~\ref{tab:kappa_human_gpt_turbo} presents the kappa agreement scores between human judgment and \gptfturbo{} across six models, as measured on the complete datasets of \etestbed{} and \htestbed{}. We notice that, overall, the agreement scores are consistently high (>0.8) and exhibit minimal variation across different models and evaluation sets. It is interesting to note that the agreement is consistently higher for well-performing models (\mixtral{} and \chatgpt{}), while the evaluation conducted by \gptf{} becomes more challenging when judging the output of poorly performing models.

\begin{table}[!htp]
    \begin{center}
    
    \resizebox{\columnwidth}{!}{
\begin{tabular}{lcccc}

\toprule

& \multicolumn{2}{c}{\textit{\gptfturbo}} & 
\multicolumn{2}{c}{\textit{\texttt{Human}}} \\ 

& \footnotesize \etestbed{} &  \footnotesize \htestbed{} 
& \footnotesize \etestbed{} &  \footnotesize \htestbed{} \\

\midrule
\multicolumn{5}{c}{\textit{Finetuning Models}} \\
\midrule

\flan{-base} & 91\% & 88\% &  90\% & 88\% \\
\godel{-base}  & 88\% & 88\% & 88\% & 89\% \\
\llama{} & 89\% & 92\% & 91\% &  90\% \\

\midrule
\multicolumn{5}{c}{\textit{3-shot Models}} \\
\midrule

\llama{} &  90\% & 89\% &  90\% &  90\% \\
\mixtral{} & 93\% & 93\% & 95\% & 94\% \\
\chatgpt{} & 93\% & 94\% & 95\% & 95\% \\

\bottomrule

\end{tabular}

    }	
 \end{center}	
	
\caption{Kappa agreement scores  between \gptf{} and \gptfturbo{} judgments (first two columns), and between \gptf{} and human judgments (last two columns). This experiment was conducted on a randomly selected subset of 110 examples from both \etestbed{} and \htestbed{}, comprising 6 models.}
\label{tab:kappa_human_gpt_gpt_turbo}

\end{table}

To ensure the quality of our evaluation, we measured the discrepancy between \gptf{} and \gptfturbo{} judgments by comparing their outputs on a random sample of 110 examples, which constitutes approximately 10\% of the total data in \testbed{}. Table~\ref{tab:kappa_human_gpt_gpt_turbo} shows the agreement scores of \gptf{}, not only with \gptfturbo{} (first two columns) but also with human judgments (last 2 columns). This comparison is based on selected 110 subset examples from \etestbed{} and \htestbed{}. On one hand, we observe a relatively high agreement between \gptf{} and \gptfturbo{}, ranging from 0.8 at worst to 0.94 at best. Notably, most disagreements occur with the less performing models (\godel{-base} and \flan{-base}), which are in line with the observations made in Table~\ref{tab:kappa_human_gpt_turbo}. Although not directly comparable\footnote{We measured the kappa agreement between \gptfturbo{} and human judgments on the randomly selected example subsets and found that the agreement strongly aligns with that observed in the full evaluation sets, with a maximum variance of $\pm0.1$ and $\pm0.2$ in rare cases.}, we notice that \gptf{} judgments are systematically closer to human ones compared to those of \gptfturbo{} across different settings. For instance, the agreement between \gptf{} and human judgments on \chatgpt{} \htestbed{} is higher by 0.4-0.5 than that of \gptfturbo{} with humans (0.91). Despite this, we believe that \gptfturbo{} presents an acceptable quality-cost trade-off, being three times less expensive than \gptf{}. By all means of comparison, it offers a comprehensively superior alternative to FaithDial's automatic evaluation metrics.

\begin{figure*}
    \centering
    \begin{subfigure}[b]{1.0\textwidth}
    \includegraphics[width=1.0\textwidth,height=0.45\textheight]{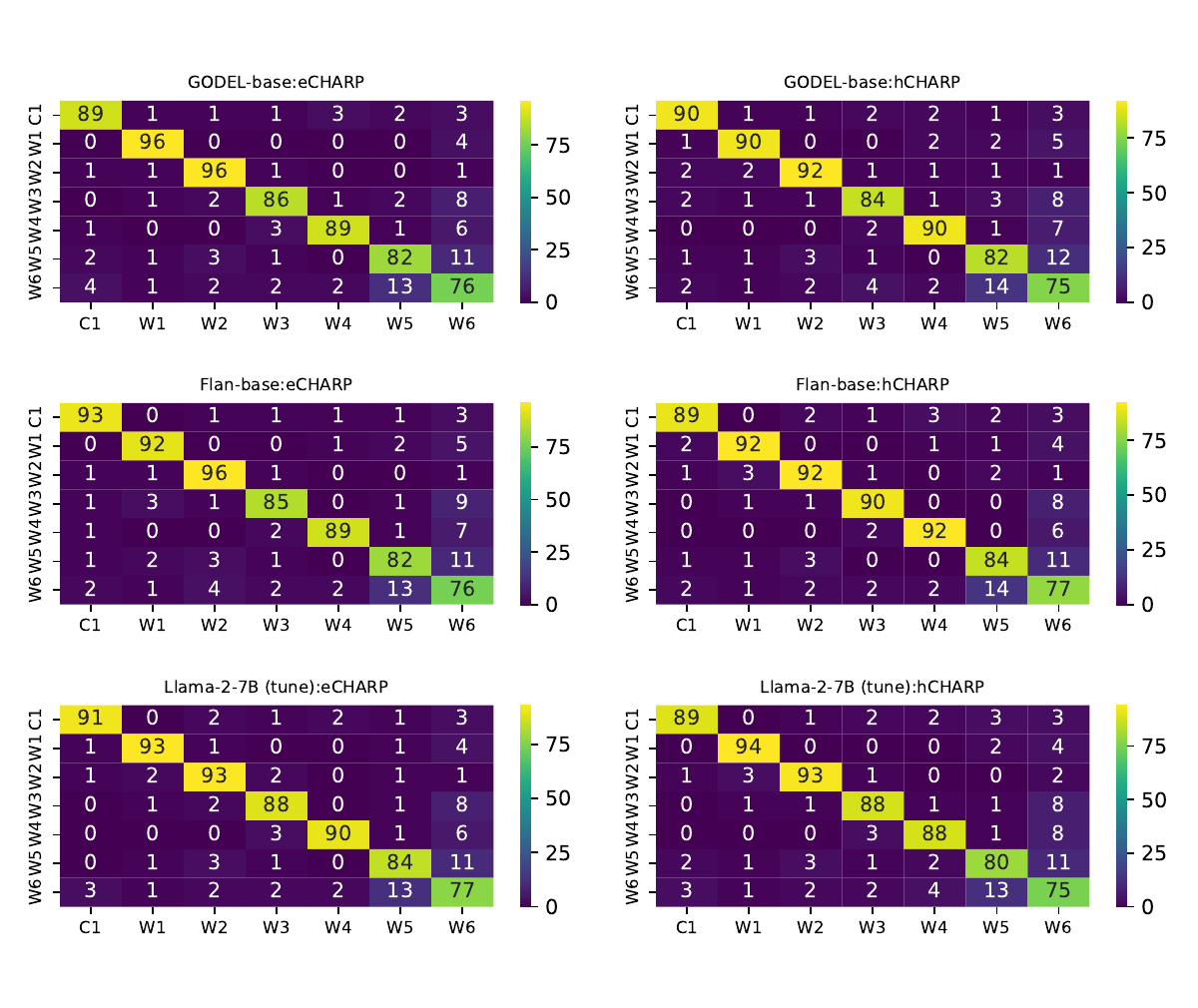}
    \end{subfigure}

    \begin{subfigure}[b]{1.0\textwidth}
       \includegraphics[width=1.0\textwidth,height=0.45\textheight]{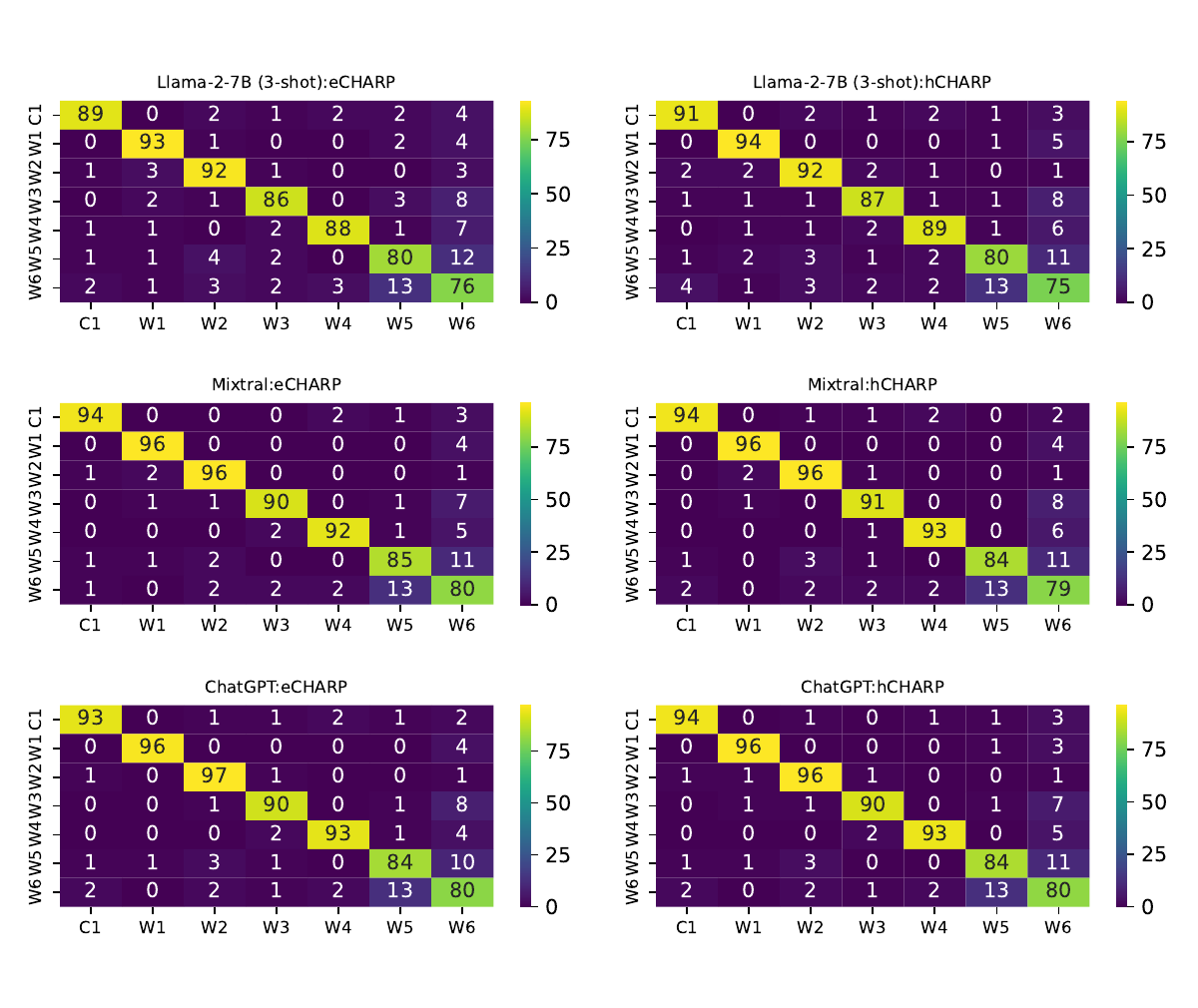}
    \end{subfigure}

    \caption{Heatmap showing the normalized (percentage) contingency tables of evaluation categories between \gptfturbo{} (rows) and human (columns) judgments. It was measured on the output of 6 models for both \etestbed{} (on the left) and \htestbed{} (on the right).}
    \label{fig:heatmap_app}
\end{figure*}

\begin{figure*}[!ht]
    \centering
    \begin{subfigure}[b]{\textwidth}
       \includegraphics[width=\textwidth]{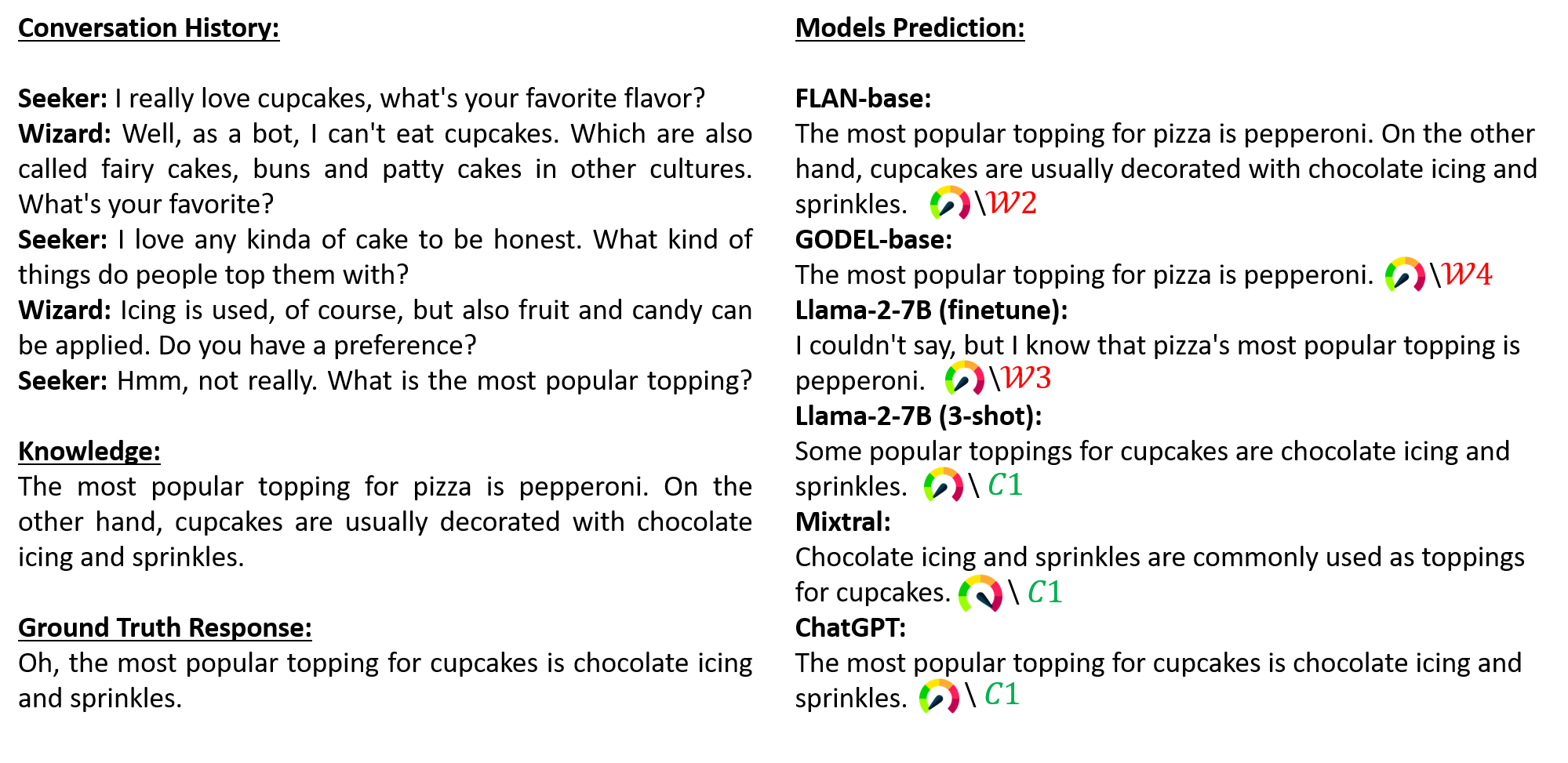}
    \end{subfigure}
    
    \begin{subfigure}[b]{\textwidth}
       \includegraphics[width=1\textwidth]{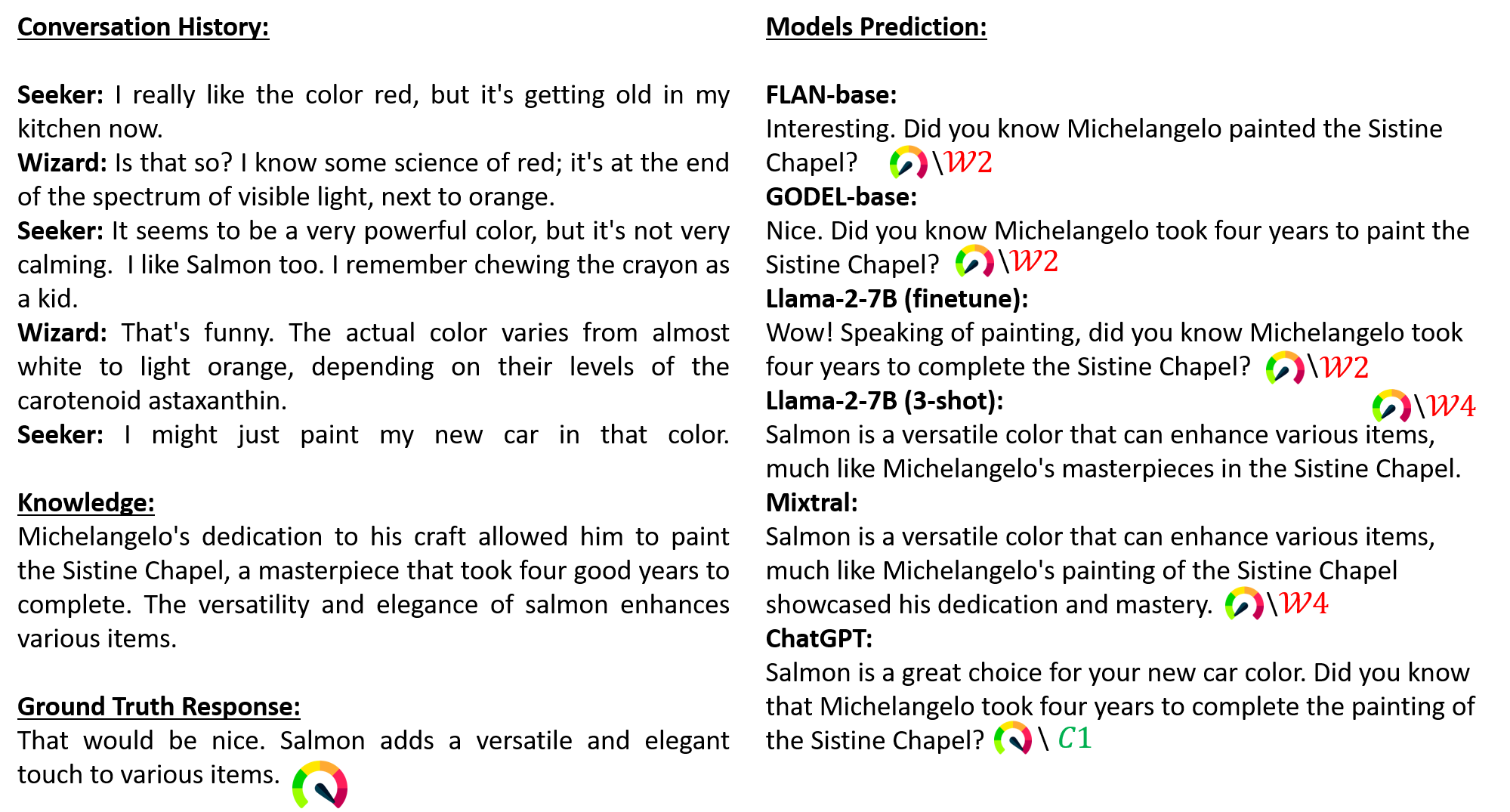}
     \end{subfigure}

    \caption{Tow examples from \htestbed{} (left side), along with the predictions of the six models employed in our study (right side). For each model response, we show the FaithDial judgment (hallucination indicated by \halIcon{}, and no hallucination by \nohalIcon{}), along with the category of human judgment. In the second example, \chatgpt{}'s response (rare but interesting) is deemed correct by human evaluators because it accurately addresses the user's comment before introducing an unrelated piece of knowledge in a manner that opens a new topic, Although it aligns with FaithDial guidelines, but the \critic{} judge this case as hallucination, mainly because \textit{painting new car} is not mentioned in the provided knowledge. }
    \label{fig:eval_example}
\end{figure*}

\end{document}